\documentclass[10pt,twocolumn,letterpaper]{article}

\usepackage{cvpr}              

\definecolor{cvprblue}{rgb}{0.21,0.49,0.74}
\usepackage[pagebackref,breaklinks,colorlinks,allcolors=cvprblue]{hyperref}
\usepackage[accsupp]{axessibility}  

\def\paperID{2} 
\def\confName{CVPR}
\def\confYear{2025}

\title{Efficient Task-specific Conditional Diffusion Policies: \\ Shortcut Model Acceleration and SO(3) Optimization}

\author{Haiyong Yu \quad Yanqiong Jin \quad Yonghao He \quad Wei Sui \\
D-Robotics\\
Beijing, China\\
{\tt\small \{joe.yu, yanqiong.jin, yonghao01.he, wei.sui\}@d-robotics.cc}
}

\begin{document}
\maketitle
\begin{abstract}
Imitation learning, particularly Diffusion Policies based methods, has recently gained significant traction in embodied AI as a powerful approach to action policy generation. These models efficiently generate action policies by learning to predict noise. However, conventional Diffusion Policy methods rely on iterative denoising, leading to inefficient inference and slow response times, which hinder real-time robot control. To address these limitations, we propose a Classifier-Free Shortcut Diffusion Policy ($\textbf{CF-SDP}$) that integrates classifier-free guidance with shortcut-based acceleration, enabling efficient task-specific action generation while significantly improving inference speed. Furthermore, we extend diffusion modeling to the $SO(3)$ manifold in shortcut model, defining the forward and reverse processes in its tangent space with an isotropic Gaussian distribution. This ensures stable and accurate rotational estimation, enhancing the effectiveness of diffusion-based control. Our approach achieves nearly 5× acceleration in diffusion inference compared to DDIM-based Diffusion Policy while maintaining task performance. Evaluations both on the RoboTwin simulation platform and real-world scenarios across various tasks demonstrate the superiority of our method.
\end{abstract}    
\section{Introduction}
\label{sec:intro}

In recent years, imitation learning \cite{hussein2017imitation} in embodied AI has emerged as one of the key directions in robotics research. The generative method based on diffusion models \cite{sohl2015deep} \cite{song2019generative} \cite{song2020improved} \cite{song2020score} \cite{ho2020denoising} \cite{song2020denoising}(i.e., Diffusion Policy \cite{chi2023diffusion} \cite{janner2022planning}) has received extensive attention for its powerful generative ability and outstanding performance in high-dimensional continuous control tasks.

Specifically, In embodied AI, robots navigate dynamic environments while performing tasks like grasping and placing. Traditional methods using hand-crafted features \cite{liu2024ok} or reinforcement learning \cite{thrun2000reinforcement} struggle with generalization and data efficiency. Diffusion policy leverages generative models to learn adaptable action strategies, enhancing robustness and decision-making.

Diffusion policies play a crucial role in the realm of embodied intelligence. For instance, in robotic manipulation tasks \cite{argall2009survey} \cite{levine2016end}, they can effectively mimic and learn expert strategies \cite{hussein2017imitation}, extracting general operational patterns from human demonstrations. Moreover, they can be integrated with visual perception \cite{cheng2024yolo} \cite{he2024light} and language understanding modules, enabling robots to execute natural language commands in open environments and boosting their interactive capabilities. To efficiently deploy embodied intelligence strategies in real-world applications, high-quality data collection \cite{fu2024mobile} \cite{chi2024universal} \cite{wang2024dexcap} \cite{kofman2005teleoperation} and task modeling are essential, a challenge that recent research has made significant strides toward addressing.

Despite their impressive capabilities, diffusion policies face challenges in data efficiency, real-time inference, and action representation. Addressing these issues is crucial for improving their practicality and effectiveness.

{\bf 1) Efficient Task-Specific Data Utilization:} Diffusion models require extensive high-quality data, but collecting such data \cite{o2024open} in robotic manipulation \cite{brohan2022rt} \cite{brohan2023rt}is costly and constrained. While multi-task data enhances generalization, it introduces challenges in handling diverse tasks with varying objectives, constraints, and action scales. Developing a unified policy representation to capture these complexities remains a key challenge.

{\bf 2) Real-Time Diffusion Inference:} Conventional diffusion models require extensive denoising iterations, making real-time robotic control impractical. Recent studies \cite{wang2024one} \cite{lipman2022flow} \cite{lu2024simplifying} aim to reduce steps while preserving action quality. However, faster denoising demands capturing more information per step; failure leads to blurry, discontinuous, or abnormal actions, compromising control robustness. Ensuring stability and quality with fewer iterations remains a challenge.

{\bf 3) Rational Action Space Representation:} Common rotation representations such as Euler angles, quaternions, and matrices each have limitations: singularities \cite{hemingway2018perspectives}, double-cover issues \cite{kuipers1999quaternions}, and redundancy \cite{evans2001rotations}. Designing a representation suitable for diffusion policies while ensuring continuity and stability is challenging. A proper rotation representation and loss function are essential to accurately capture errors and maintain balance with translation and other parameters, preventing optimization imbalances.

To address precision and real-time challenges in task-specific imitation learning with diffusion based methods, this paper introduces a unified, efficient conditional diffusion policy head for task-specific action generation. We incorporate a classifier-free conditional modeling approach into the diffusion policy, allowing the model to learn action generation directly from task-specific expert data. The model leverages conditional information, such as task labels, environmental states, or model features to guide the diffusion process. This strategy helps capture both the commonalities and unique characteristics across different tasks. To boost the efficiency of the diffusion inference process, we integrate a shortcut model into the conditional diffusion framework. This approach introduces the number of iterations as an condition during model diffusion, enabling high-quality policy generation in just one or several steps. Consequently, the model can recover action sequences with fewer iterations, significantly accelerating the denoising process while maintaining the coherence and accuracy. Recognizing the non-Euclidean nature \cite{wolfe2012introduction} of rotations in robotic control, we model the rotation representation on the 3D rotation Lie group $SO(3)$ \cite{chevalley2018theory} and design a loss function tailored to these manifold characteristics. This allows the model to capture and recover rotational components more reasonable during action generation, thereby enhancing the expressiveness of the generated actions.

The contributions of this paper are summarized as follows.
\begin{itemize}
\item We introduce a Classifier-Free Shortcut Diffusion Policy ($\textbf{CF-SDP}$) to learn task-specific action generation while ensuring high control performance. Moreover, by incorporating the Shortcut Model, the diffusion denoising process can enhance the inference speed, making real-time robot control achievable.

\item We extend diffusion modeling to the $SO(3)$ manifold in the shortcut diffusion policy, formulating the forward and reverse processes in its tangent space with an isotropic Gaussian distribution. This ensures more accurate and stable rotational estimation, improving the overall effectiveness of diffusion-based control policies.

\item We conducted comparative experiments across various tasks in both RoboTwin \cite{mu2024robotwin} simulation environments and real scenarios. Compared with the DDIM of Diffusion Policy, the classifier-free conditional shortcut diffusion policy we proposed achieves nearly a 5x acceleration in diffusion denoising inference efficiency, while maintaining its performance without degradation.
\end{itemize}

This paper is organized as follows. In Section~\ref{sec:related_work}, we will give a brief introduction of the related works. Then, in Section~\ref{sec:method}, detailed introduction of the proposed method is presented. And in Section~\ref{sec:experiments}, experimental results and discussions are included. Finally, the conclusions are made in Section~\ref{sec:concludion}.
\section{Related Works}
\label{sec:related_work}
\subsection{Diffusion Model}
Diffusion models \cite{sohl2015deep} \cite{song2019generative} \cite{song2020improved} \cite{song2020score} \cite{ho2020denoising} \cite{song2020denoising}, as a type of generative model \cite{goodfellow2020generative} \cite{pinheiro2021variational} \cite{lipman2022flow}, have swiftly garnered extensive attention in both the academic and industrial sectors. The fundamental concept of diffusion models was initially put forward by Sohl-Dickstein \cite{sohl2015deep}. It involves transforming a complex data distribution into a simple Gaussian noise distribution by gradually adding noise, and then restoring the original data structure through learning the reverse denoising process. Building upon this idea, DDPM \cite{ho2020denoising} was proposed, which enabled diffusion models to achieve remarkable achievements in tasks such as image generation. Subsequently, approaches like DDIM \cite{song2020denoising} and those based on score matching \cite{song2020score} further enhanced the sampling efficiency and generation speed, laying the groundwork for the application of diffusion models in real-time tasks. Latent Diffusion Models \cite{rombach2022high} conduct diffusion modeling by mapping high-dimensional images into a low-dimensional latent space. This significantly cuts down on computational costs while preserving the crucial semantic information of the images. The GLIDE model \cite{nichol2021glide} combines conditional diffusion with classifier guidance, also focuses on text-to-image generation and image editing tasks. GLIDE excels in generating realistic and creative images, demonstrating the formidable power of diffusion models when handling multi-modal inputs, and offering a new research direction for conditional image generation. In image generation tasks, conditional diffusion models have been extensively studied to produce images that meet specific semantic requirements. Dhariwal and Nichol et al. proposed the Classifier Guidance method \cite{dhariwal2021diffusion}. By leveraging a pre-trained classifier to provide gradient guidance during the reverse denoising process, this method makes the generation process more compliant with the conditional requirements. This approach has demonstrated outstanding performance in text-to-image generation tasks \cite{shuai2024survey}, but it also increases the complexity of the model, as it relies on an external classifier network.

\subsection{Diffusion Policy}

Diffusion models have drawn attention to generate high-quality, multi-modal samples. In robotic control, the introduction of Diffusion Policy \cite{chi2023diffusion} has brought a fresh perspective to action generation. By framing robot action generation as a conditional denoising problem and leveraging the step-by-step reverse generation process of diffusion models, Diffusion Policy captures multi-modal distributions in high-dimensional action spaces, enabling stable modeling of complex vision-motion mappings. Researchers have further explored diffusion strategies in 3D scenes, known as 3D Diffusion Policy \cite{ze20243d}. This approach aims to directly generate high-quality 3D action trajectories using 3D scene information \cite{qi2017pointnet++}, tackling tasks that involve intricate spatial geometries. Meanwhile, Flow Matching \cite{lipman2022flow} proposes an alternative to the traditional diffusion process by learning continuous flow fields through optimal transport trajectories, thereby simplifying model training and boosting sampling efficiency. Consistency Policy \cite{prasad2024consistency} achieves one-step or few-step fast generation by mapping from arbitrarily noised directly to the original data. However, the distillation and pretraining process can be complex, especially when training from scratch, as it requires carefully designed time step schedules and loss functions. Additionally, auto-regressive generation \cite{brohan2022rt} \cite{brohan2023rt} strategies produce actions sequentially rather than generating an entire sequence in one go, which enhances accuracy and stability in robotic control tasks.

Building on these advances, we propose a unified and efficient classifier-free \cite{ho2022classifier} conditional diffusion policy head. Furthermore, we introduce a Shortcut Model \cite{frans2024one} into this conditional diffusion framework to accelerate the denoising process.

\subsection{Task-Specific and Hierarchical Policy Learning}

Task-Specific and hierarchical learning are fundamental approaches to addressing complex challenges in robotic control. 
The Hierarchical Diffusion Policy \cite{ma2024hierarchical} disentangles task planning from low-level action generation. This design not only strengthens the ability to generate multi-task strategies but also enhances adaptability to complex task variations. RDT \cite{liu2024rdt} combines diffusion models with Transformer to improve generalization, leveraging self-attention to capture long-range dependencies for superior performance in multi-task scenarios. $\pi$\_0 \cite{black2024pi_0}, a universal framework based on Decision Diffusion, facilitates efficient decision-making in multi-task robotic operations by utilizing cross-task representation learning, enabling seamless knowledge transfer and improved generalization. Meanwhile, 3D Diffuser Actor \cite{ke20243d} is a diffusion-based policy tailored for 3D robotic manipulation \cite{qi2017pointnet++}, modeling the diffusion process directly within a 3D action space to enhance both the precision and stability of robotic execution.

These methods rely on intricate multi-modal models or task planning mechanisms to extract expert data representations across tasks. We propose a unified and efficient conditional diffusion policy head designed to significantly enhance task-specific action generation.

\subsection{Rotation Space Representation}

Diffusion policies based on DDPM \cite{ho2020denoising} traditionally operate in Euclidean space. However, conventional rotation representations in Euclidean space suffer from singularities and redundancy issues: Euler angles exhibit singularities \cite{hemingway2018perspectives}, quaternions have a double-cover property \cite{kuipers1999quaternions}, and rotation matrices are highly redundant \cite{evans2001rotations}. To address these limitations, researchers have extended diffusion policies to the $SO(3)$ manifold \cite{chevalley2018theory}. Leach et al. \cite{leach2022denoising} proposed a probabilistic denoising diffusion model on $SO(3)$, focusing on rotational alignment, effectively handling rotational uncertainty and improving alignment accuracy. Taking this further, Jagvaral et al. \cite{jagvaral2024unified} introduced a unified diffusion generative model framework explicitly designed for $SO(3)$ space, successfully applying it to computer vision and astrophysics. Meanwhile, Grasp Diffusion Network \cite{carvalho2024grasp} leverages diffusion models within the $SO(3)$ space to generate grasping strategies, extracting effective grasping information from partial point clouds, demonstrating the importance of jointly modeling rotational and translational information for enhanced grasp accuracy.

While these approaches integrate $SO(3)$ into the DDIM, our method takes a step further by implementing rotational diffusion on the $SO(3)$ manifold within the shortcut model-based diffusion training and inference process. By using an isotropic Gaussian distribution to handle rotational variations, our approach could capture the intrinsic patterns of robot pose changes in complex tasks.
\section{Method}
\label{sec:method}

\subsection{Problem Formulation}

Given an expert data sampled from the real data distribution ${\bf x}_0 \sim q({\bf x})$, we define a forward diffusion process that incrementally corrupts the data by adding Gaussian noise over $T$ steps. This process generates a sequence of noisier samples ${\bf x}_1, \dots, {\bf x}_T$. where the noise level at each step is regulated by a variance schedule ${\beta_t \in (0, 1)}_{t=1}^T$, ensuring a smooth transition from clean data to pure noise. As the $t$ increases, the data samples transforms its distictive distribution into noise.
\begin{equation}
    q({\bf x}_{1:T} \vert {\bf x}_{0}) = \prod^T_{t=1} q({\bf x}_t \vert {\bf x}_{t-1})
    \label{eq:1}
\end{equation}
\begin{equation}
    q({\bf x}_t \vert {\bf x}_{t-1}) = \mathcal{N}({\bf x}_{t}; \sqrt{1 - \beta_t}{\bf x}_{t-1}, \beta_t{\bf I})
    \label{eq:2}
\end{equation}

In reverse diffusion process, we can reconstruct the sample from a gaussian noise input ${\bf x}_T \sim q({\bf 0}, {\bf I})$ to estimate $q({\bf x}_{t-1} \vert {\bf x}_{t})$. Estimating $q({\bf x}_{t-1} \vert {\bf x}_{t})$ requires access to the entire dataset, which is challenging. We should train a model $p_\theta({\bf x}_{t-1} \vert {\bf x}_{t})$ to approximate the conditional probabilities.
\begin{equation}
    p_\theta({\bf x}_{0:T}) = p({\bf x}_T)\prod^T_{t=1} p_\theta({\bf x}_{t-1} \vert {\bf x}_{t})
    \label{eq:3}
\end{equation}
\begin{equation}
    p_\theta({\bf x}_{t-1} \vert {\bf x}_{t}) = \mathcal{N}({\bf x}_{t-1};{\bf \mu}_\theta({\bf x}_t, t), {\bf \Sigma}_\theta({\bf x}_t, t))
    \label{eq:4}
\end{equation}
where ${\bf \mu}_\theta({\bf x}_t, t)$ is means and ${\bf \Sigma}_\theta({\bf x}_t, t)$ is variance.

The objective of diffusion models is to train a model to estimate ${\bf \mu}_\theta({\bf x}_t, t)$ and ${\bf \Sigma}_\theta({\bf x}_t, t)$, maximizing the log-likelihood of the prediction distribution. Empirically, Ho et al. discovered that using a simplified objective to train the diffusion model to predict sample noise ${\bf \epsilon}^t$ yields better results.
\begin{equation}
    {\bf L} = \nabla \| {\bf \epsilon}^t - {\bf \epsilon}_\theta({\bf x}_0, {\bf \epsilon}^t, t) \|
    \label{eq:5}
\end{equation}

In task-specific robotic control, our goal is to develop a unified policy capable of generating task-specific actions across diverse scenarios. We consider an task-specific expert dataset ${\bf D} = {({\bf O}_i, {\bf A}_i, {\bf C}_i)}^N_{i=1}$, where $N$ denotes the total number of distinct tasks in the dataset, ${\bf O}_i$ denotes the observation, ${\bf A}_i$ denotes the corresponding action, ${\bf C}_i$ denotes the task-specific condition such as task identifiers, task descriptions, or language instructions. Our objective is to model the conditional distribution $p({\bf A} \vert {\bf O}, {\bf C})$ that can effectively translate high-dimensional observation and arbitrary task conditions cues into robust action sequences.

Traditional diffusion-based methods such as 3D Diffusion Policy frame action generation as a reverse diffusion process, gradually transforming Gaussian noise into structured actions. However, these methods struggle with effectively capturing task-specific variations in a multi-task setting. Moreover, existing methods typically operate in euclidean space and require a large number of denoising steps, limiting their efficiency for real-time applications. To address these challenges, we propose a Classifier-Free Shortcut Diffusion Policy with $SO(3)$-based rotational representation, enhancing efficiency and precision in task-specific robotic action generation.

\begin{figure}[ht]
    \centering
    \includegraphics[width=0.45\textwidth]{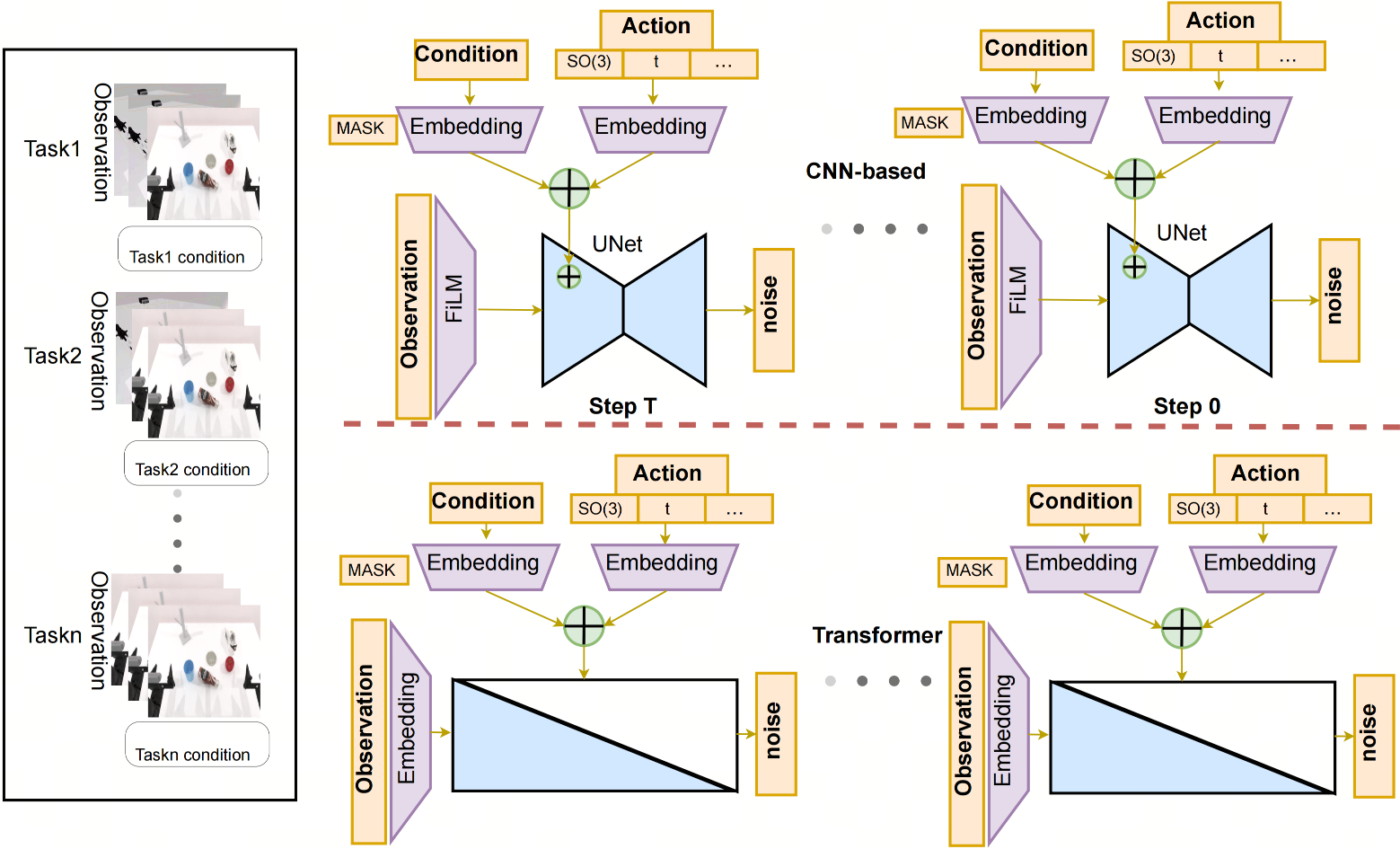}
    \caption{Overview of Classifier-Free Shortcut Diffusion Policy with SO(3)-based rotational representation}
    \label{fig:overview}
\end{figure}

An overview of our method is presented in Fig.~\ref{fig:overview}, which is implemented using both CNN-based and Transformer-based architectures. Each model takes observations, task conditions, and actions as inputs to predict the corresponding noise, with the rotation in the action represented using $SO(3)$.

\subsection{Classifier-free Shortcut Diffusion Policy}

To overcome the limitations of conventional diffusion policies, we propose a Classifier-Free Shortcut Diffusion Policy ($\textbf{CF-SDP}$) that integrates classifier-free guidance \cite{ho2022classifier} and shortcut-based acceleration \cite{frans2024one} to both enhance task-specific conditioning and accelerate the denoising process. The proposed method is built upon two fundamental principles that serve as the foundation for its design and effectiveness.

First, we adopt shortcut policy to learn robotic visuomotor policies, which enables the generation of high-quality actions in just one or very few sampling steps, significantly accelerating the inference process.

Specifically, for each task, we define the forward diffusion process that interpolates noisy action data ${\bf A}^t_{i}$.
\begin{equation}
    {\bf A}^t_{i} = (1-t){\bf A}_0 + t{\bf A}_1
    \label{eq:6}
\end{equation}
where ${\bf A}_0 \sim \mathcal{N}(0, {\bf I})$ denotes a noise data distribution and ${\bf A}_1 \sim \mathcal{D}$ denotes the real action data distribution.

The velocity ${\bf v}^t_i$ is defined as the direction from the noise data to the real action data.
\begin{equation}
    {\bf v}^t_{i} = {\bf A}_1 - {\bf A}_0
    \label{eq:7}
\end{equation}

The key intuition is that model support different sampling budgets by conditioning both the timestep $t$ and a desired step size $d$. To condition on step size $d$, the shortcut $s({\bf A}^t_i,{\bf O}^t_i,t,d)$ is designed as normalized direction from ${\bf A}^t_i$ to next point ${\bf A}^{t+d}_i$.
\begin{equation}
    {\bf A}^{t+d}_i = {\bf A}^t_i + s({\bf A}^t_i,{\bf O}^t_i,t,d)d
    \label{eq:8}
\end{equation}

The shortcut model has an inherent self-consistency property:
\begin{equation}
    \begin{aligned}
    s({\bf A}^t_i,{\bf O}^t_i,t, 2d) &= s({\bf A}^t_i,{\bf O}^t_i,t,d)/2 \\
                                     &+ s({\bf A}^{t+d}_i,{\bf O}^t_i,t,d)/2
    \end{aligned}
    \label{eq:9}
\end{equation}

Thus, we train a parameter model $s_\theta({\bf A}^t_i,{\bf O}^t_i,t,d)$ to estimate the shortcut model $s({\bf A}^t_i,{\bf O}^t_i,t,d)$.

The training loss objective learns a mapping from noise to data that remains consistent when queried across any sequence of step sizes:
\begin{equation}
\begin{aligned}
    {\bf L} &= {\bf MSE}(s_\theta({\bf A}^t_i,{\bf O}^t_i,t, 0), {\bf v}^t_{i}) + {\bf MSE}(s_\theta({\bf A}^t_i,{\bf O}^t_i,t, 2d), \\
            &\hspace{1.2em} s_\theta({\bf A}^t_i,{\bf O}^t_i,t,d)/2 + s_\theta({\bf A}^{t+d}_i,{\bf O}^t_i,t,d)/2)
    \label{eq:10}
\end{aligned}
\end{equation}

Second, we incorporate classifier-free guidance into the shortcut policy, enabling direct integration of task conditions to effectively control the task-specific diffusion generation process.

Under the premise of DDIM, We refer to a conditional diffusion model where the reverse process is conditioned on task-specific information according to :
\begin{equation}
    \begin{split}
    p_\theta({\bf A}^{t-1}\vert{\bf A}^{t}, {\bf C}^{t}) &\sim \mathcal{N}({\bf A}^{t-1}; \mu({\bf A}^{t}, t) \\
    &+\gamma\sigma^2 * \nabla log{p_\theta}({\bf C}^{t}\vert{\bf A}^{t}), \sigma^2{\bf I})
    \end{split}
    \label{eq:11}
\end{equation}

To capture the conditional distribution $p({\bf A}\vert{\bf O},{\bf C})$ in classifier-free shortcut diffusion policy, we integrate the representation of conditions into the actions and modify the Equation.~\eqref{eq:11} accordingly and retrain the model.
\begin{equation}
    \begin{aligned}
        {\bf L} &= {\bf MSE}(s_\theta({\bf A}^t_i\oplus{\bf C}^t_i,{\bf O}^t_i,t, 0), {\bf v}^t_{i}) \\
        &+ {\bf MSE}(s_\theta({\bf A}^t_i\oplus{\bf C}^t_i,{\bf O}^t_i,t, 2d), s_\theta({\bf A}^t_i\oplus{\bf C}^t_i,{\bf O}^t_it,d)/2 \\
        &+ s_\theta({\bf A}^{t+d}_i\oplus{\bf C}^{t+d}_i,{\bf O}^t_i,t,d)/2)
        \label{eq:12}
    \end{aligned}
\end{equation}

\subsection{Shortcut Diffusion Policy in $SO(3)$}

Instead of performing diffusion in euclidean space, we model the rotation component directly in the special orthogonal group $SO(3)$. Utilizing the classiﬁer-free shortcut diffusion policy mechanism, we define the forward and reverse process on the tangent space of $SO(3)$ with an isotropic Gaussian distribution \cite{berman1980isotropic}, enabling efficient and accurate rotation estimation.

We represent a 6d pose of action ${\bf G}=({\bf \xi})$ as an element of the $se(3)$ Lie algebra, where $\xi \in \mathbb{R}^6$ is the twist coordinate consisting of a translation component $v \in \mathbb{R}^3$ and a rotation component $\omega \in \mathbb{R}^3$, where the rotation part can be interpreted as the axis-angle representation.

Specifically, we define the rotation vector ${\bf r}$ as the Lie algebra representation of the rotation part in $se(3)$ which belongs to $so(3)$. Here ${\bf r}=(r_1, r_2, r_3) \in \mathbb{R}^3$, with each $r_i$ constrained to the range $(0, \pi)$. Subsequently, the rotation vector ${\bf r}$ is converted into $SO(3)$ matrix ${\bf R \in \mathbb{R}^{3 \times 3}}$ through the exponential map. We also define the translation part $ \mathbf{s} $ as a vector in $ \mathbb{R}^3 $, where $ \mathbf{s} = (s_1, s_2, s_3) $ represents the translational components along the $ x $, $ y $, and $ z $ axes respectively.

Based on these preliminary definitions, we can rewrite the forward and reverse processes of the classiﬁer-free shortcut diffusion policy model.

In forward process, the noise addition process for the rotation component requires performing interpolation in $SO(3)$, as shown in Equation.~\eqref{eq:13}. This is necessary because rotations are elements of a non-Euclidean space, and the noise needs to be added in a way that respects the manifold structure of $SO(3)$. While the noise addition process for the translation component is straightforward, as shown in Equation.~\eqref{eq:14}.
\begin{equation}
    {\bf R}^t_i = \exp(t\log{\bf R}^1_i + (1-t)\log{\bf R}^0_i)
    \label{eq:13}
\end{equation}
\begin{equation}
    {\bf s}^t_i = t{\bf s}^1_i + (1-t){\bf s}^0_i
    \label{eq:14}
\end{equation}

In the reverse process, we can again express the classifier-free shortcut diffusion policy model for both the rotational and translational components in the case of $SO(3)$ for rotations and $\mathbb{R}^3$ for translations.

The calculation of the translation part ${\bf s}$ is straightforward. Let's expand and rewrite the Equation.~\eqref{eq:10}.
\begin{equation}
    \begin{aligned}
        {\bf L}_s = &\| s_\theta({\bf s}^t_i,{\bf O}^t_i,{\bf C}^t_i,t, 0) - ({\bf s}^1_{i} - {\bf s}^0_{i}) \|^2 + \\
                  &\|s_\theta({\bf s}^t_i,{\bf O}^t_i,{\bf C}^t_i,t, 2d) - (s_\theta({\bf s}^t_i,{\bf O}^t_i,{\bf C}^t_i,t,d)/2 \\
                  &+ s_\theta({\bf s}^{t+d}_i,{\bf O}^t_i,{\bf C}^t_i,t,d)/2) \|^2
        \label{eq:15}
    \end{aligned}
\end{equation}
where $ {\bf s}^{t+d}_i =  {\bf s}^t_i + s_\theta({\bf s}^t_i,{\bf O}^t_i,{\bf C}^t_i,t,d)d $.

The rotation part is handled through the application of $SO(3)$ interpolation, which involves computing the transformation from one rotation state to the next. This process can be described by the following equation:
\begin{equation}
    \begin{aligned}
        {\bf L}_R &= \| s_\theta({\bf R}^t_i,t, 0) - \log({\bf R}^1_{i} {\bf R}^0_{i}) \|^2 + \|s_\theta({\bf R}^t_i,t, 2d) \\
                &- (s_\theta({\bf R}^t_i,t,d)/2 + s_\theta({\bf R}^{t+d}_i,t,d)/2) \|^2
        \label{eq:16}
    \end{aligned}
\end{equation}

where
\begin{equation}
    \log{\bf R}^{t+d}_i =  \log{\bf R}^t_i + s_\theta({\bf R}^t_i,{\bf O}^t_i,{\bf C}^t_i,t,d)d
    \label{eq:17}
\end{equation}

so
\begin{equation}
    {\bf R}^{t+d}_i =  {\bf R}^t_i \exp(s_\theta({\bf R}^t_i,{\bf O}^t_i,{\bf C}^t_i,t,d)d)
    \label{eq:18}
\end{equation}

The two objectives ${\bf L}_s$ and ${\bf L}_R$ are combined which guides the network to learn smooth and consistent rotational transformations, effectively mitigating issues related to singularities and redundancy that are common in Euclidean representations.

\subsection{Design Details Options}

\textbf{Classifier-Free Network Architecture:} Based on the theoretical derivation of the Classifier-Free Shortcut Diffusion Policy, we explored three different model architectures to incorporate task-specific conditions:
(1) Adding condition features directly in the action space.
(2) Integrating condition features into the observation space.
(3) Using FiLM model \cite{perez2018film} to fuse condition features in the observation space.
All three approaches enable the model to leverage conditions to guide task-specific action generation effectively.

\textbf{Shortcut Model with CNN and Transformer Architectures:} For the shortcut model, we implemented both CNN-based \cite{chi2023diffusion} and Transformer-based architectures \cite{peebles2023scalable} tailored for Diffusion Policy and 3D Diffusion Policy tasks. In practice, the Transformer-based architecture for 3D Diffusion Policy was highly sensitive to hyperparameters, making it challenging to train effectively. As a result, we primarily focused on the CNN-based architecture for 3D Diffusion Policy, which delivered the best performance. Additionally, we tested both architectures for Diffusion Policy. With these architectural optimizations, our model inference speed achieved nearly a 5× acceleration compared to traditional DDIM models.

\textbf{End-Effector and Simulation}: In the Classifier-Free Shortcut Diffusion Policy, actions were initially represented by joint angles. But upon the integration of $SO(3)$, we shifted to using end - effector poses. The majority of our tasks were assessed within the RoboTwin simulator \cite{mu2024robotwin}. Nevertheless, as it offers limited support for controlling robotic arms based on end-effector poses, we also conducted tests of our model in other simulation environments.
\section{Experiments}
\label{sec:experiments}
The experiments are divided into three parts. First, we analyze the performance of CF-SDP across various tasks under different conditions. Second, we conduct ablation studies on a single task to validate the effectiveness of the proposed method. Third, we evaluate our method in a real robot munipulation task.

\subsection{Dataset and Hyperparameters}
All training and evaluation data for the simulation environments are collected using RoboTwin, a widely adopted mobile bimanual platform equipped with 14-degree of-freedom robots.This benchmark contains a variety of dual-arm tasks, such as pick \& place, beat blocks, and handover.

For real-world robot data collections, we utilize the 6-DoF RealMan arms along with a head-mounted camera. This setup composes a 14-dimensional action space, with data recorded at 15Hz. A RealSense L515 camera is used to capture point cloud data.

For all tasks, we utilize the AdamW optimizer with a learning rate of $1*10^{-4}$ and a weight decay of $0.1$. The batch size is set to 256. For each training mini-batch, we use $k=1/4$ to split the data into two parts, one for calculate flow-matching target and the other for self-consistency target. The model is trained on an NVIDIA 4090 GPU with 24GB memory, with a maximum of $3K$ epochs.

\subsection{Experiments on Specific Task Conditional}

This section presents the results of the experiments on specific tasks with different conditions. In order to show the ability to control action generation through specific task conditions, a task scenario is firstly designed using RoboTwin. We placed the objects required for the three different tasks in the same scene, and added two other objects that are not related to any of the tasks to increase the richness of the scene, see Fig.\ref{fig:multi_task}. The descriptions of the three different tasks are as follows:
\begin{figure}[htbp]
  \centering
  \includegraphics[width=0.4\textwidth]{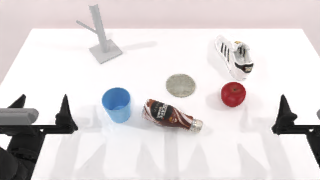}
  \caption{specific task conditions experimental scene setup.}
  \label{fig:multi_task}
\end{figure}

\textbf{A. bottle adjust:} This task is to pick up a bottle horizontally on the table and put it upright. The bottle is placed on the table with a random rotation.

\textbf{B. pick apple:} This task is to pick up an apple and lift it. The apple is placed randomly on the table.

\textbf{C. empty cup place:} The robotic arm places an empty cup on the cup mat. The empty cup and the cup mat are randomly placed on the table.

Note that the positions of objects irrelevant to the three different tasks are also randomly placed on the table. As our approach aims to verify the specific task conditions guidance ability of the model, we use the same task scenario as described before to collect training data and evaluate the model. We collected 100 demonstrations and set specific labels for each task, resulting in a total of 300 demonstrations. The specific task conditions are used to guide the model in generating actions that are consistent with the task conditions. The conditions here can be any text that can uniquely represent the corresponding task, such as the task name, task condition, even digitals, etc. Without loss of generality, we use integers to describe the conditions of a specific task. For example, we use 0,1,2 to represent bottle adjust task, pick apple task, and empty cup place task, respectively.

\begin{figure}[htbp]
  \centering
  \includegraphics[width=0.4\textwidth]{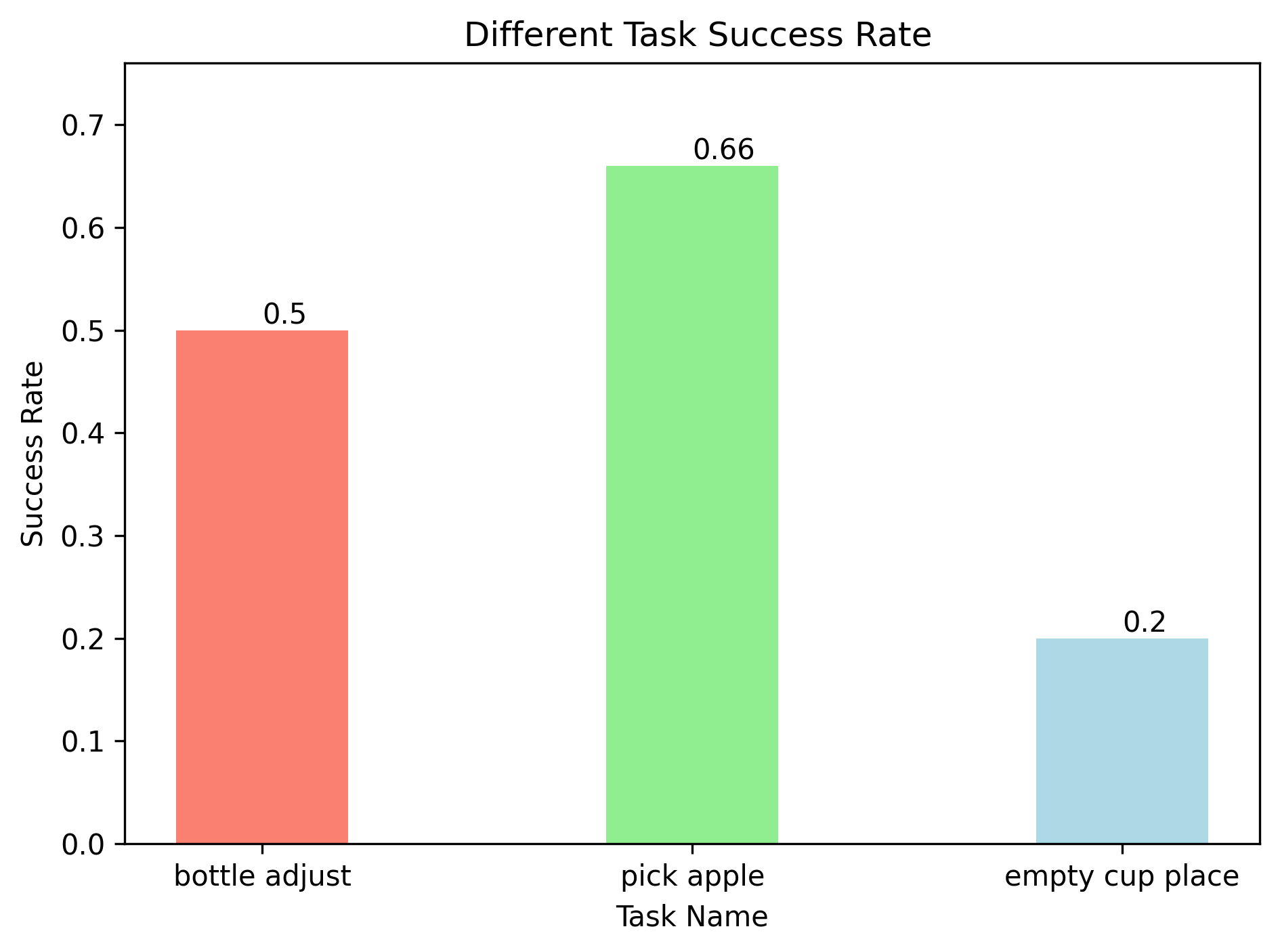}
  \caption{CF-SDP success rate on different tasks with task-Conditional(results for 100 demonstrations).}
  \label{fig:CF-SDP_single}
\end{figure}

The experimental results are shown in Fig.\ref{fig:CF-SDP_single}, where our method can work well on all three tasks in this scenario. This result shows that robotic arms action generation can be guided by specific task conditions. 

To futher demonstrate the performance of our method, the 3D Diffusion Policy (DP3) is also compared with the CF-SDP. The DP3 model is modified to a model that can support guidance by specific task conditions, and trained on the same dataset as CF-SDP. We use 20,50,100 samples to train DP3 model and CF-SDP model. For fair comparison, we set 10 denoising steps for DP3 and CF-SDP. As illustration in Fig, the CF-SDP can achieve better performance than DP3 on almost all three tasks.
\begin{table}[h]
  \centering
  \renewcommand{\arraystretch}{1.}
  \setlength{\tabcolsep}{5.0pt}
  \resizebox{\columnwidth}{!}{
    \tiny
    \begin{tabular}{lccc|lccc}
        \toprule
        \multicolumn{1}{l}{Task} & \multicolumn{3}{c|}{DP3 with Classifier-Free Guidance} & \multicolumn{1}{l}{Task} & \multicolumn{3}{c}{CF-SDP} \\
        \cmidrule(lr){2-4} \cmidrule(lr){6-8}
        & 20 & 50 & 100 &  & 20 & 50 & 100 \\
        \midrule
        bottle adjust & 33.33$\pm$7.97 & 32.33$\pm$1.25 & 43.33$\pm$2.62 & bottle adjust & \textbf{34.67$\pm$3.40} & \textbf{40.33$\pm$4.99} & \textbf{44.0$\pm$5.89} \\
        pick apple & 20.0$\pm$1.63 & 35.67$\pm$4.03 & 48.67$\pm$4.19 & pick apple & \textbf{36.33$\pm$3.68} & \textbf{59.33$\pm$1.70} & \textbf{72.33$\pm$6.94} \\
        empty cup place & 11.67$\pm$7.32 & \textbf{26.0$\pm$1.63} & \textbf{37.0$\pm$11.52} & empty cup place & \textbf{13.0$\pm$4.32} & 21.67$\pm$4.50 & 21.0$\pm$6.48 \\
        \bottomrule
    \end{tabular}
  }
  \caption{Comparison of success rate between DP3 with Classifier-Free Guidance and CF-SDP under task-Conditional(results for 20, 50 and 100 demonstrations).}
  \label{tab:multi_task_performance}
\end{table}

\subsection{Experiments on Simulation Environments}

We present a series of experiments evaluating efficiency and precision of our approach in simulation environments. Based on RoboTwin, we carefully compare our approach to DP3, by training SDP(CF-SDP without CF) with datasets of 50 and 100 samples. In all experiments, we use a standard depth image resolution of $320\times 180$ to get point cloud, collected by the RealSense L515 camera.

Following the training setup in DP3, the policy was trained 3 seeds for each test with seed number 0, 1, 2. For each policy, we evaluate 100 eposides with 3000 training epochs and then calculate mean and standard deviation of the three success rates, which used to report below. For shortcut diffusion policy, we evaluate models on samples generated with 10, 5, 3, 2 and 1 denoising diffison step(s).

\begin{figure}[htbp]
  \centering
  \includegraphics[width=0.5\textwidth]{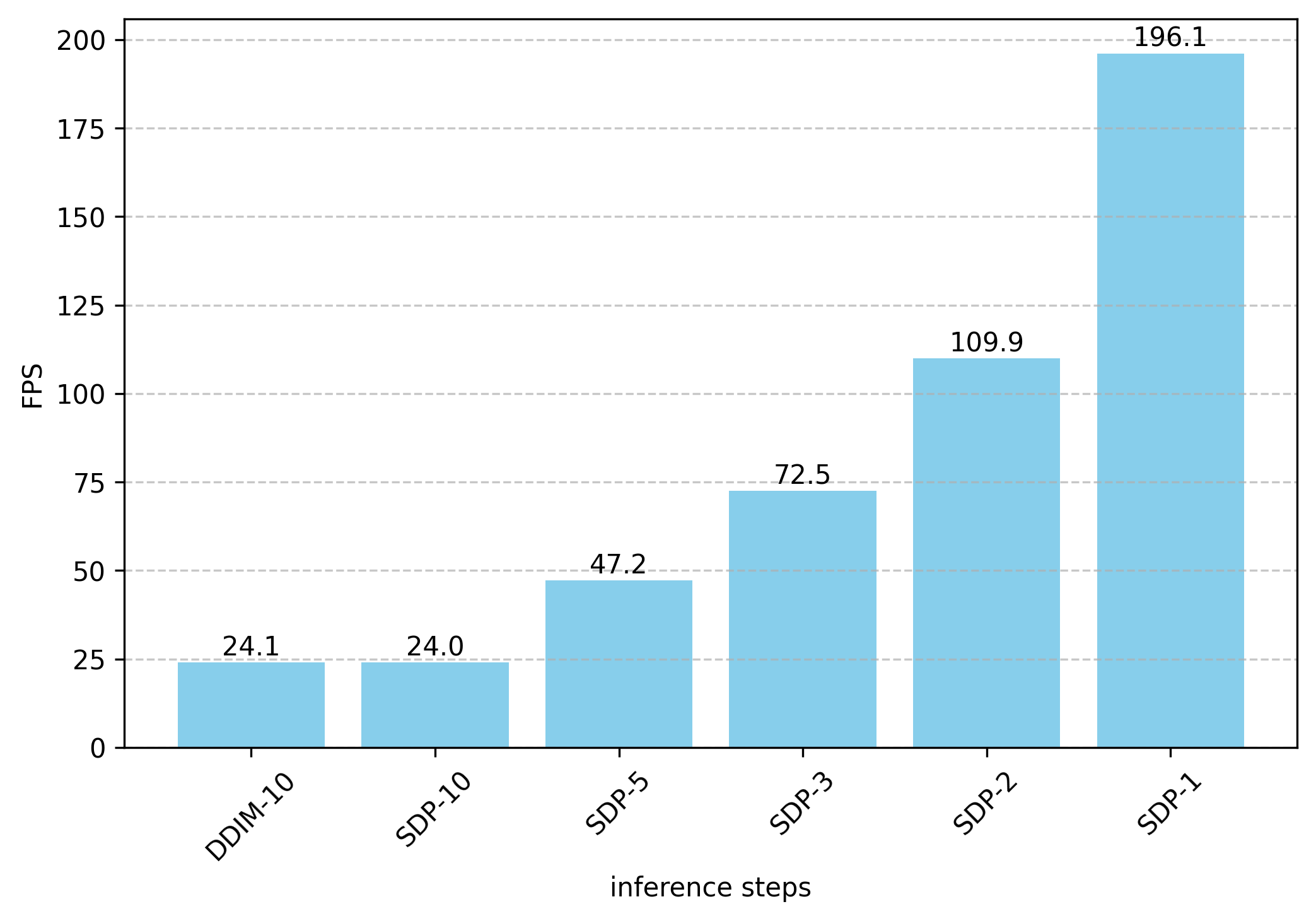}
  \caption{DP3 vs. SDP: inference speed comparison, SDP denoising with 10, 5, 3, 2, 1 denoising diffusion step(s).}
  \label{fig:fps_comparison}
\end{figure}

As depicted in Fig.\ref{fig:fps_comparison}, the SDP model achives an inference speed marginally surpassing DP3 with 10 denoising steps. However, as the number of inference steps decreases, the inference speed is significantly improved, so that the model can meet higher real-time requirements. To this end, we need to further explore whether using fewer reasoning steps can maintain good reasoning accuracy. 

It is worth mentioning that, our approach can achieve similar performance on few-step generation. The experimental results are presented in Table.\ref{tab:performance}.
Baseline results, including the 3D Diffusion Policy, is reported by RoboTwin. Notably, across all tasks and diverse settings, our proposed SDP achieves the better average success rate, even with 3 or 2 denoising diffison steps. In addition, we observe that the SDP with one-step denoising diffusion, can achieve a success rate comparable to DP3 on more than $90\%$ of the tasks.

In terms of efficiency, the shortcut diffusion policy show few-step even one-step generation capability competetive with DP3, which used DDIM with 10 denoising steps. At the same time, when the shortcut diffusion policy models use the same denoising steps as DP3, better performance can be achieved in almost all tasks. Otherwise, the SDP model can maintain performance of base diffusion models on few-step generation.

\begin{table}[h]
  \centering
  \renewcommand{\arraystretch}{1.0}
  \setlength{\tabcolsep}{5.0pt}
  \resizebox{\columnwidth}{!}{
    \tiny
    \begin{tabular}{lcc|lcc}
        \toprule
        \textbf{Task} & 50 & 100 & \textbf{Task} & 50 & 100 \\
        \midrule
        \multicolumn{3}{l|}{\textit{Block Hammer Beat}} & \multicolumn{3}{l}{\textit{Block Handover}} \\
        \midrule
        DP3(DDIM-10) & 58.3$\pm$6.5 & 49.7$\pm$8.1 & DP3(DDIM-10) & 85.0$\pm$15.6 & 63.7$\pm$7.0 \\
        SDP-10 & 76.0$\pm$12.9 & \textbf{81.0$\pm$5.72} & SDP-10 & 72.7$\pm$11.26 & 82.3$\pm$2.06 \\
        SDP-5 & 77.0$\pm$10.98 & 75.3$\pm$9.98 & SDP-5 & 77.3$\pm$11.9 & 83$\pm$5.35 \\
        SDP-3 & 76.0$\pm$12.33 & 80.3$\pm$8.18 & SDP-3 & 83.7$\pm$9.74 & 81.7$\pm$2.87 \\
        SDP-2 & \textbf{77.3$\pm$6.94} & 78.7$\pm$7.54 & SDP-2 & \textbf{88.7$\pm$3.40} & 84$\pm$5.72 \\
        SDP-1 & 41.7$\pm$7.73 & 62.3$\pm$6.2 & SDP-1 & 88.0$\pm$5.10 & \textbf{89.3$\pm$4.19} \\
        \midrule
        \multicolumn{3}{l|}{\textit{Blocks Stack (Easy)}} & \multicolumn{3}{l}{\textit{Blocks Stack (Hard)}} \\
        \midrule
        DP3(DDIM-10) & 17.0$\pm$7.0 & 22.0$\pm$1.0 & DP3(DDIM-10) & 1.7$\pm$0.6 & 3.0$\pm$1.7 \\
        SDP-10 & 17.0$\pm$2.45 & \textbf{25.3$\pm$4.78} & SDP-10 & \textbf{3.67$\pm$0.94} & \textbf{3.33$\pm$0.94} \\
        SDP-5 & \textbf{19.7$\pm$3.40} & 24.7$\pm$1.25 & SDP-5 & 2.67$\pm$0.47 & 3.0$\pm$0.82 \\
        SDP-3 & 18.3$\pm$1.7 & 24.3$\pm$2.87 & SDP-3 & 2.0$\pm$0.36 & 2.33$\pm$0.47 \\
        SDP-2 & 19.0$\pm$3.56 & 25.0$\pm$2.83 & SDP-2 & 2.0$\pm$0.82 & 1.67$\pm$0.94 \\
        SDP-1 & 12.7$\pm$2.87 & 22.3$\pm$2.62 & SDP-1 & 1.33$\pm$0.47 & 0.33$\pm$0.47 \\
        \midrule
        \multicolumn{3}{l|}{\textit{Bottle Adjust}} & \multicolumn{3}{l}{\textit{Container Place}} \\
        \midrule
        DP3(DDIM-10) & 70.7$\pm$2.5 & 72.7$\pm$10.10 & DP3(DDIM-10) & 74.0$\pm$5.6 & 89.0$\pm$7.5 \\
        SDP-10 & \textbf{70.67$\pm$4.71} & 75.0$\pm$10.03 & SDP-10 & \textbf{75.7$\pm$0.47} & \textbf{91.0$\pm$4.73} \\
        SDP-5 & 69.67$\pm$4.92 & \textbf{76.0$\pm$7.2} & SDP-5 & 73.7$\pm$2.87 & 88$\pm$2.45 \\
        SDP-3 & 69.67$\pm$6.13 & \textbf{76.0$\pm$7.26} & SDP-3 & 74.3$\pm$4.92 & 80.0$\pm$0.82 \\
        SDP-2 & 70.0$\pm$8.04 & 73.0$\pm$7.48 & SDP-2 & 74.3$\pm$2.36 & 75.3$\pm$5.31 \\
        SDP-1 & 57.33$\pm$6.94 & 71$\pm$6.38 & SDP-1 & 58.3$\pm$3.77 & 70.3$\pm$5.56 \\
        \midrule
        \multicolumn{3}{l|}{\textit{Diverse Bottles Pick}} & \multicolumn{3}{l}{\textit{Dual Bottles Pick (Easy)}} \\
        \midrule
        DP3(DDIM-10) & 34.70$\pm$6.70 & 33.70$\pm$5.90 & DP3(DDIM-10) & \textbf{60.30$\pm$7.1} & \textbf{32.0$\pm$4.60} \\
        SDP-10 & 37.0$\pm$8.04 & \textbf{44.67$\pm$16.36} & SDP-10 & 58.33$\pm$4.64 & 29.33$\pm$5.56 \\
        SDP-5 & 44.33$\pm$8.38 & 39.0$\pm$15.58 & SDP-5 & 56.67$\pm$6.02 & 29.33$\pm$8.01 \\
        SDP-3 & 42.67$\pm$6.34 & 36.67$\pm$16.05 & SDP-3 & 54.33$\pm$4.03 & 30.33$\pm$4.64 \\
        SDP-2 & \textbf{45.67$\pm$7.72} & 37.33$\pm$13.28 & SDP-2 & 45.0$\pm$11.78 & 19.0$\pm$11.23 \\
        SDP-1 & 38.67$\pm$1.89 & 36.33$\pm$5.44 & SDP-1 & 31$\pm$13.44 & 22.33$\pm$8.26 \\
        \midrule
        \multicolumn{3}{l|}{\textit{Dual Bottles Pick (Hard)}} & \multicolumn{3}{l}{\textit{Dual Shoes Place}} \\
        \midrule
        DP3(DDIM-10) & 48.0$\pm$5.2 & 57.3$\pm$4.0 & DP3(DDIM-10) & 10.0$\pm$2.6 & 12.0$\pm$2.0 \\
        SDP-10 & \textbf{49.7$\pm$2.49} & 67.7$\pm$2.87 & SDP-10 & 11.67$\pm$3.3 & 23.67$\pm$2.87 \\
        SDP-5 & 46.3$\pm$3.40 & \textbf{67.7$\pm$2.62} & SDP-5 & 11.33$\pm$4.99 & 23.33$\pm$2.63 \\
        SDP-3 & 47.7$\pm$3.77 & \textbf{67.7$\pm$2.62} & SDP-3 & \textbf{12.33$\pm$3.30} & \textbf{25.0$\pm$1.63} \\
        SDP-2 & 48.7$\pm$3.09 & 65.3$\pm$2.82 & SDP-2 & 8.33$\pm$2.06 & 24.67$\pm$1.87 \\
        SDP-1 & 39$\pm$4.97 & 53.7$\pm$6.18 & SDP-1 & 11$\pm$0.816 & 20.33$\pm$1.87 \\
        \midrule
        \multicolumn{3}{l|}{\textit{Empty Cup Place}} & \multicolumn{3}{l}{\textit{Mug Hanging (Easy)}} \\
        \midrule
        DP3(DDIM-10) & 70.3$\pm$7.2 & 71.2$\pm$20.4 & DP3(DDIM-10) & \textbf{14.0$\pm$3.60} & 14.7$\pm$3.50 \\
        SDP-10 & 77.33$\pm$15.63 & 82.33$\pm$6.24 & SDP-10 & 9$\pm$3.74 & 11$\pm$1.63 \\
        SDP-5 & \textbf{85.33$\pm$11.47} & \textbf{87.33$\pm$5.73} & SDP-5 & 12$\pm$2.94 & 11.33$\pm$4.50 \\
        SDP-3 & 83.67$\pm$11.09 & 84$\pm$10.71 & SDP-3 & 12.67$\pm$5.56 & 16.33$\pm$8.96 \\
        SDP-2 & 78.67$\pm$8.18 & 84.67$\pm$5.74 & SDP-2 & 13.0$\pm$9.43 & \textbf{16.67$\pm$6.55} \\
        SDP-1 & 73$\pm$6.98 & 78.33$\pm$3.091 & SDP-1 & 9.67$\pm$3.40 & 9.0$\pm$6.38 \\
        \midrule
        \multicolumn{3}{l|}{\textit{Mug Hanging (Hard)}} & \multicolumn{3}{l}{\textit{Pick Apple Messy}} \\
        \midrule
        DP3(DDIM-10) & 11.0$\pm$6.1 & 12.7$\pm$2.3 & DP3(DDIM-10) & 10.7$\pm$4.0 & 11.7$\pm$5.5 \\
        SDP-10 & 14.0$\pm$8.52 & 18.67$\pm$12.12 & SDP-10 & 20$\pm$7.48 & 31.67$\pm$7.93 \\
        SDP-5 & 13.0$\pm$7.79 & \textbf{20.33$\pm$15.92} & SDP-5 & \textbf{20.33$\pm$7.41} & \textbf{34.67$\pm$7.13} \\
        SDP-3 & \textbf{17.67$\pm$10.87} & 14.67$\pm$10.27 & SDP-3 & 18.33$\pm$6.6 & 29.33$\pm$5.25 \\
        SDP-2 & 12.67$\pm$9.81 & 13.33$\pm$8.73 & SDP-2 & 15.0$\pm$7.26 & 31.67$\pm$4.11 \\
        SDP-1 & 5.67$\pm$1.25 & 7.33$\pm$5.56 & SDP-1 & 16.0$\pm$5.89 & 28.0$\pm$5.10 \\

        \bottomrule
    \end{tabular}
  }
  \caption{Performance comparison across different tasks and demonstrations (results for
  50 and 100 demonstrations).}
  \label{tab:performance}
\end{table}

\subsection{Real-world Experiments }

\begin{figure}[ht]
  \centering
  \begin{subfigure}[b]{0.22\textwidth}
      \includegraphics[width=\textwidth, height=3cm]{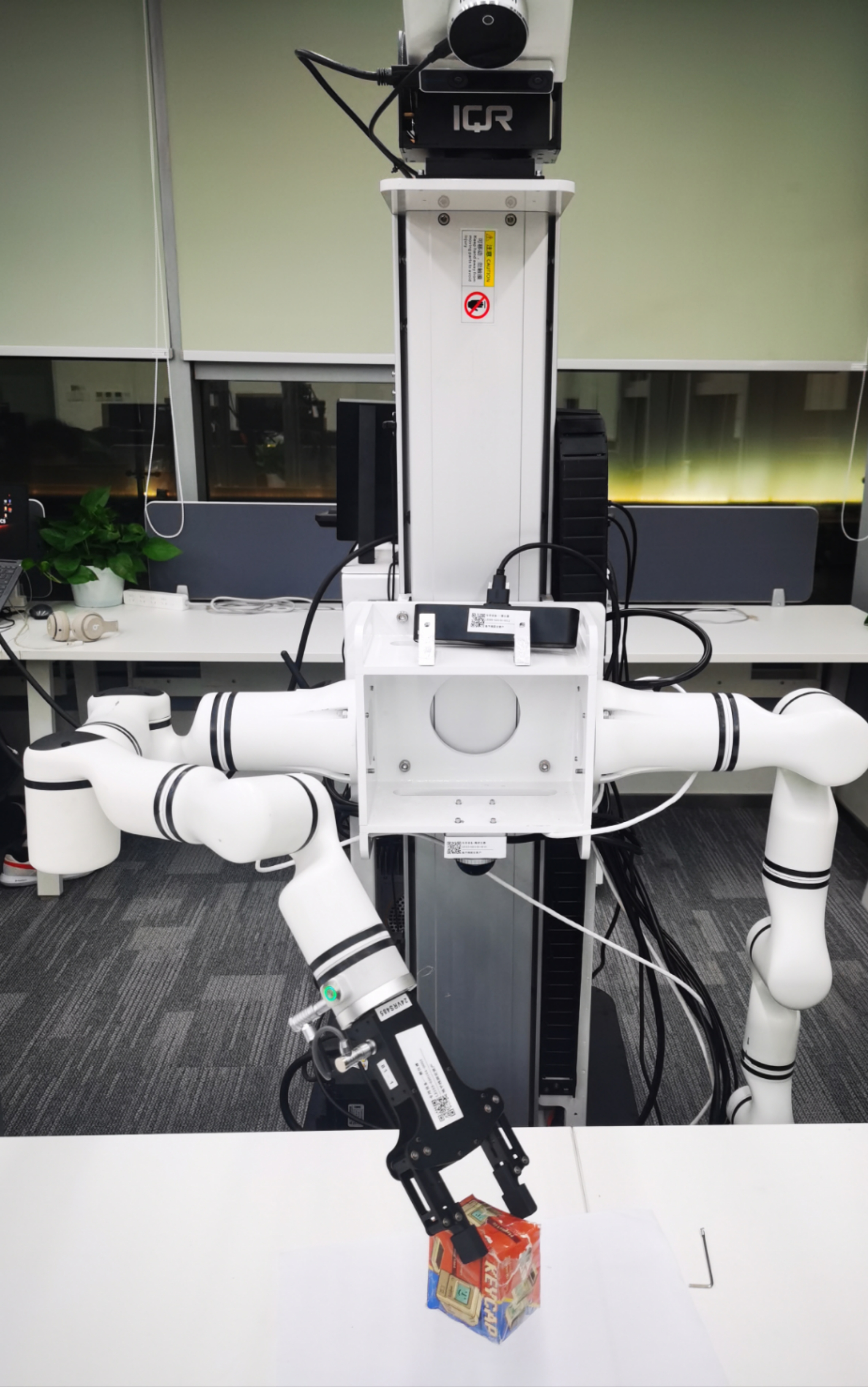}
      \caption{robotic arm and camera setup}
      \label{fig:image1}
  \end{subfigure}
  \hfill
  \begin{subfigure}[b]{0.22\textwidth}
      \includegraphics[width=\textwidth, height=3cm]{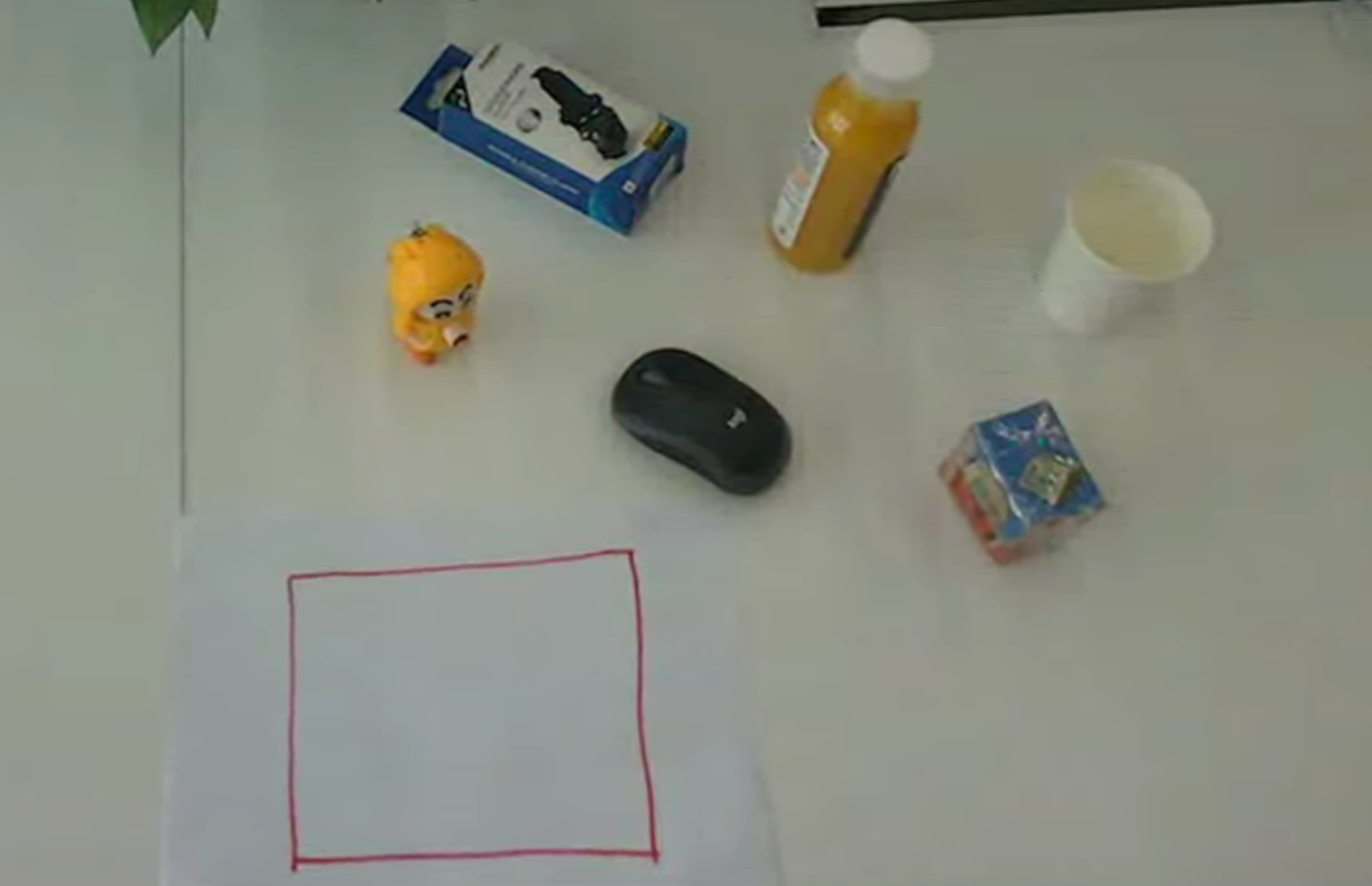}
      \caption{image from RealSense L515}
      \label{fig:image2}
  \end{subfigure}
  \caption{Real world experimental scene setup, the red box is the specific location for placing the object.}
  \label{fig:real_world_setup}
\end{figure}

In the last experiments, we evaluate how the CF-SDP generative models performance transfers from the simulation experiments to a real-world manipulation scenario. We choose five everyday objects - Box, Bottle, Cup, Toy, Mouse - to construct the task scenarios, as shown in Fig.\ref{fig:real_world_setup}. In this experiment, the RealMan arm needs to pick up an object from a table and place it at the specific location, referred to as $pnp$. A RealSense L515 camera is used to collect visual observations. Our computer is equipped with an 13th Gen Intel(R) Core(TM) i7-13700 CPU; 62 GB RAM; RTX 3090 NVIDIA GPU.

\begin{figure}[ht]
  \centering
  \begin{subfigure}[b]{0.12\textwidth}
      \includegraphics[width=\textwidth]{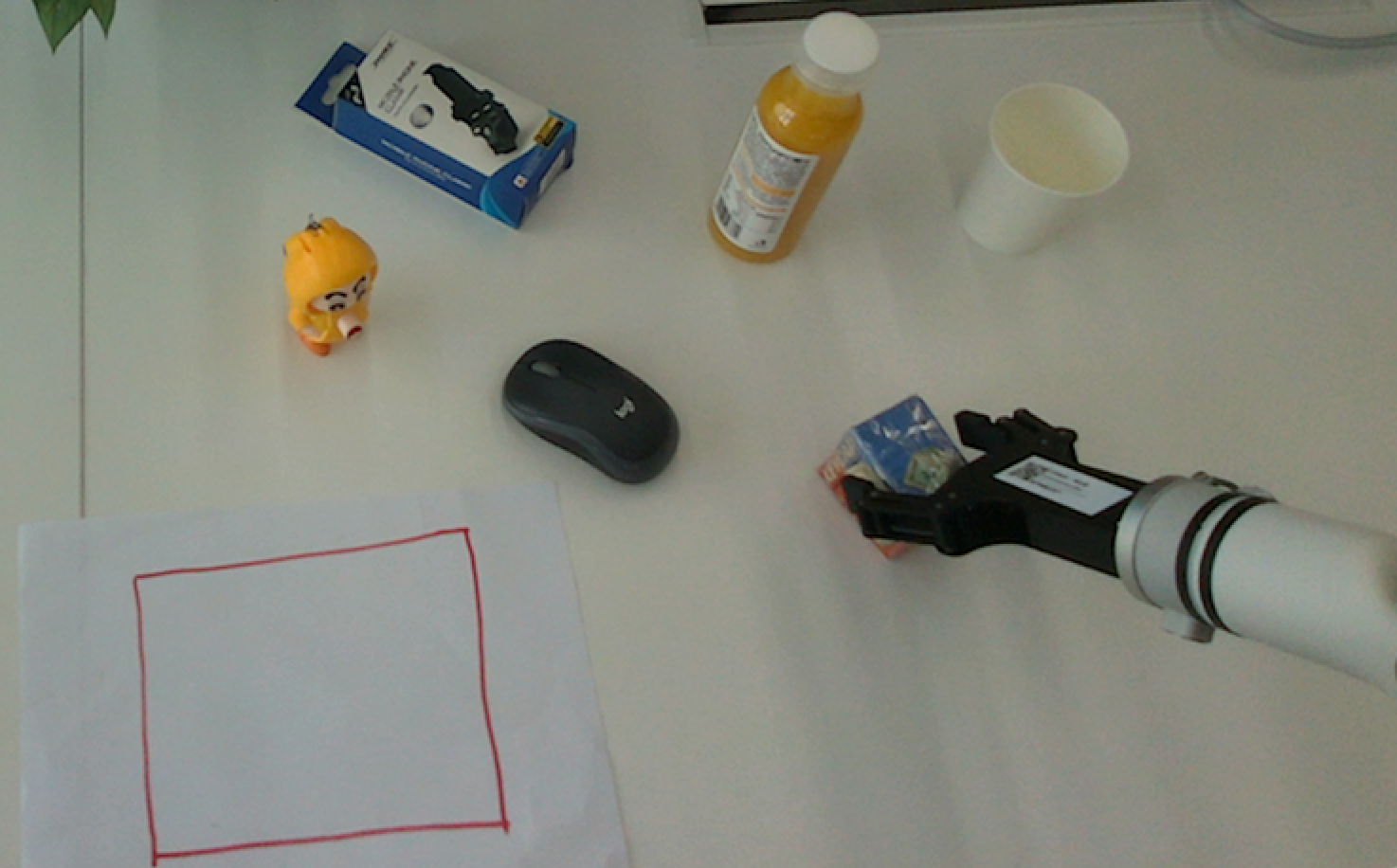}
      \label{fig:1a}
  \end{subfigure}
  ~ 
  \begin{subfigure}[b]{0.12\textwidth}
      \includegraphics[width=\textwidth]{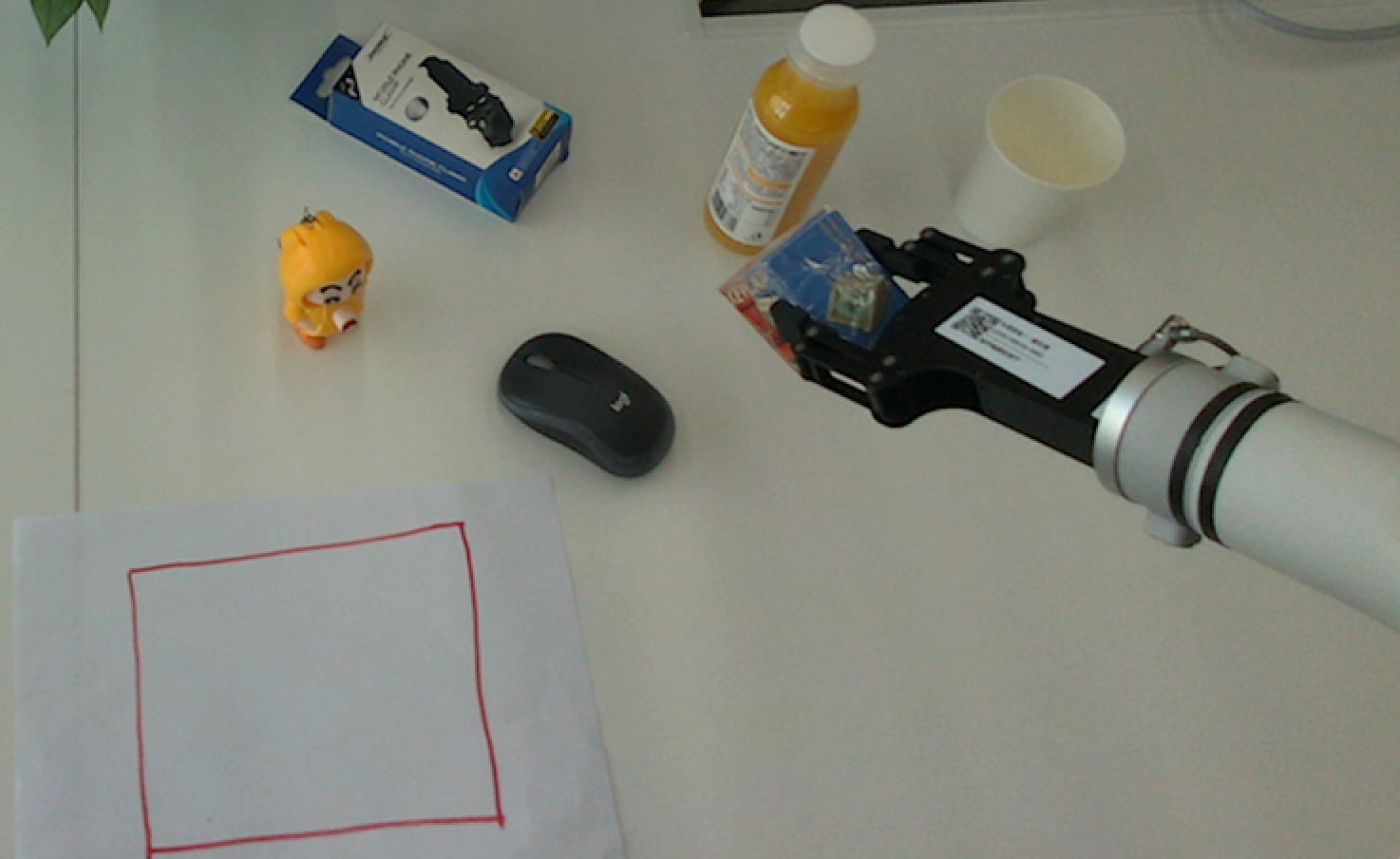}
      \label{fig:1b}
  \end{subfigure}
  ~
  \begin{subfigure}[b]{0.12\textwidth}
      \includegraphics[width=\textwidth]{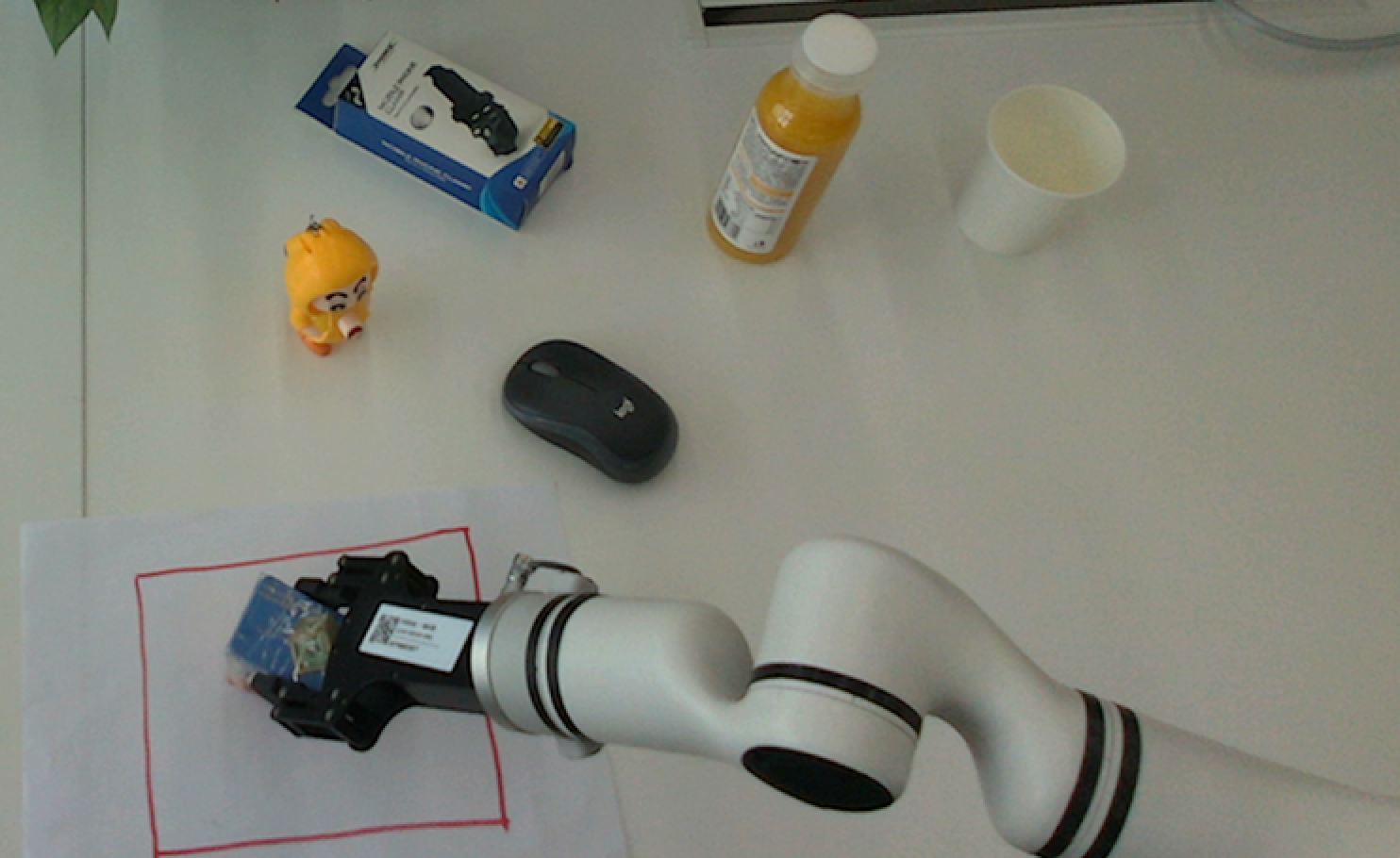}
      \label{fig:1c}
  \end{subfigure}

  \begin{subfigure}[b]{0.12\textwidth}
      \includegraphics[width=\textwidth]{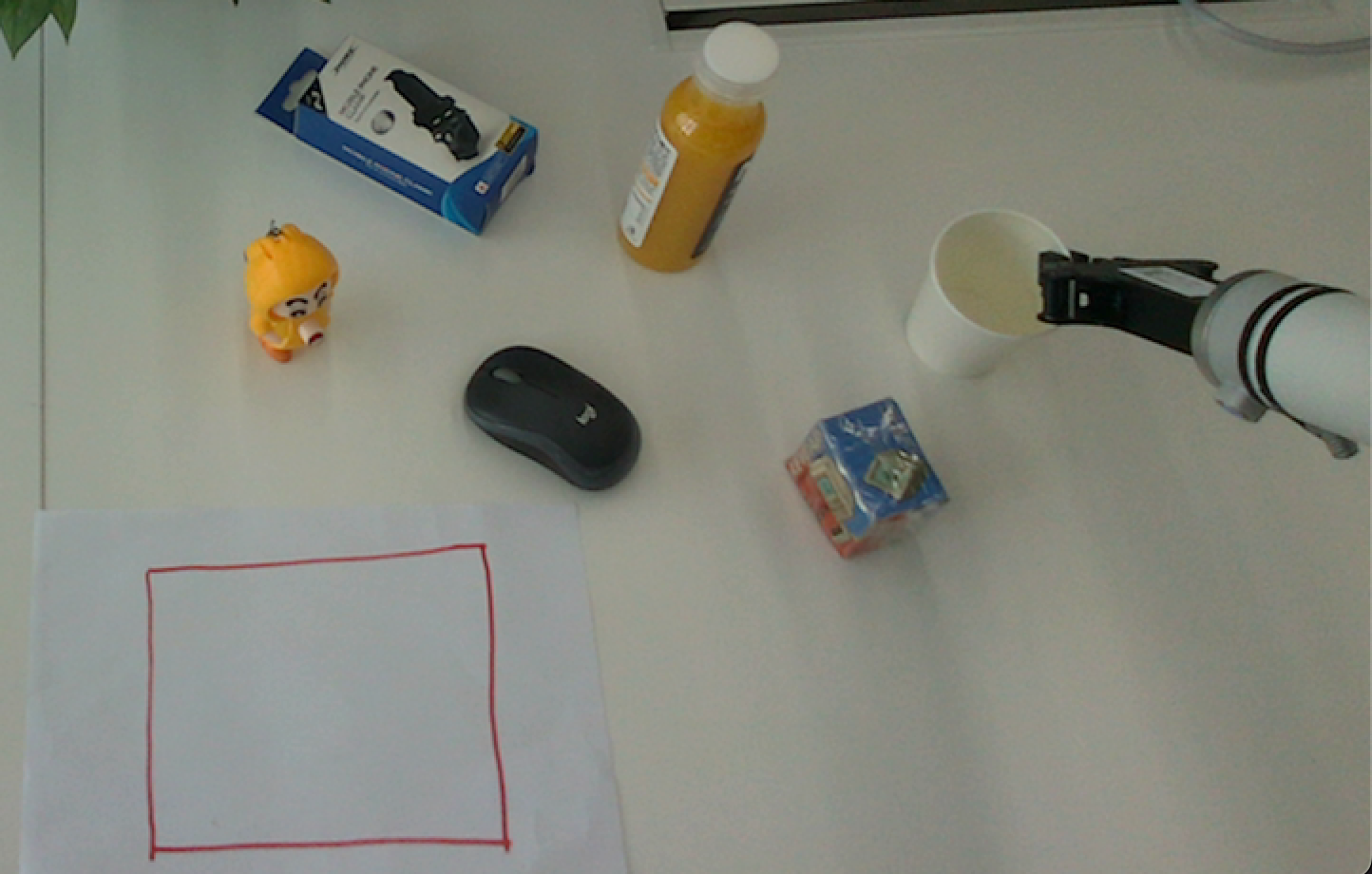}
      \label{fig:1d}
  \end{subfigure}
  ~
  \begin{subfigure}[b]{0.12\textwidth}
      \includegraphics[width=\textwidth]{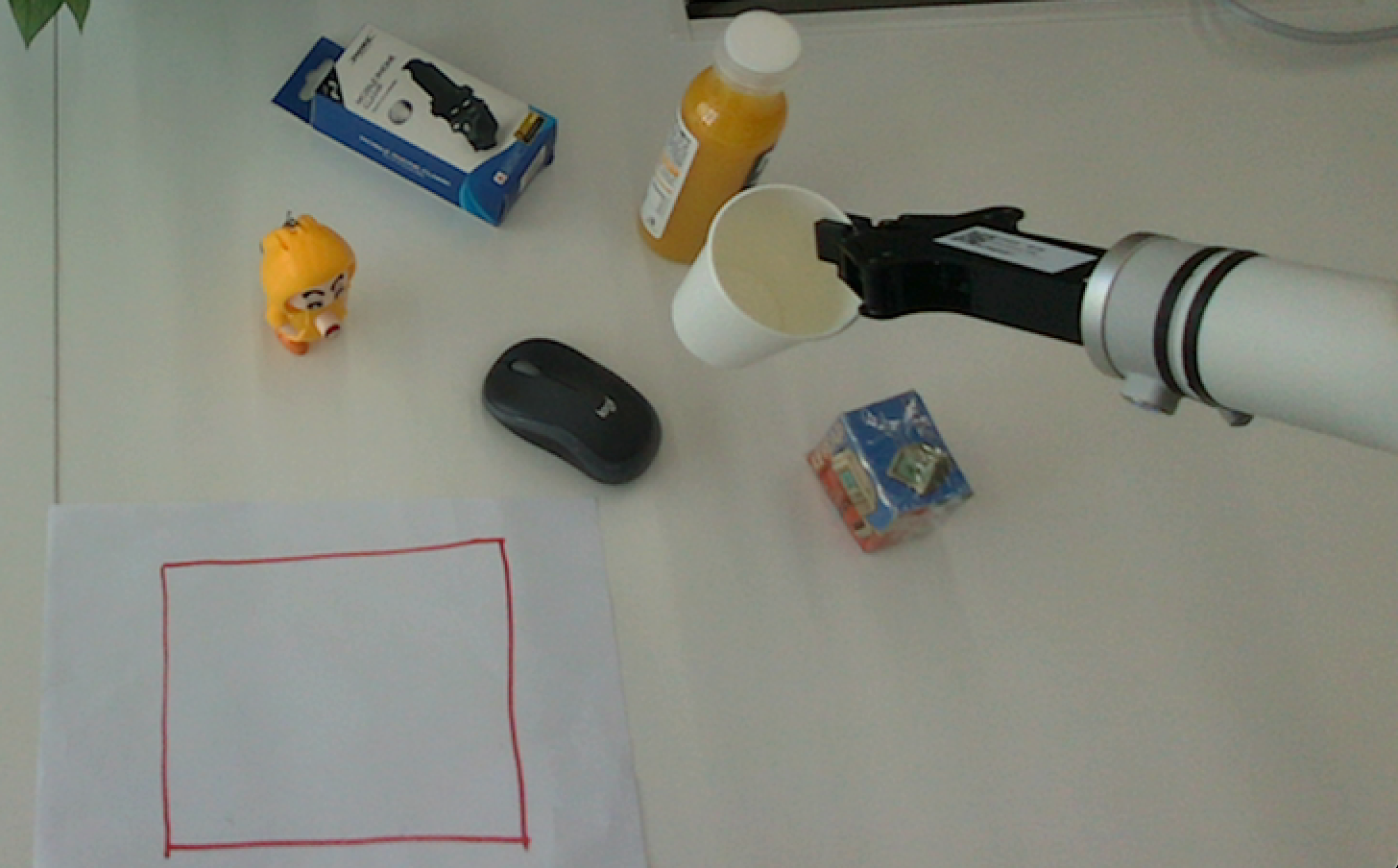}
      \label{fig:1e}
  \end{subfigure}
  ~
  \begin{subfigure}[b]{0.12\textwidth}
      \includegraphics[width=\textwidth]{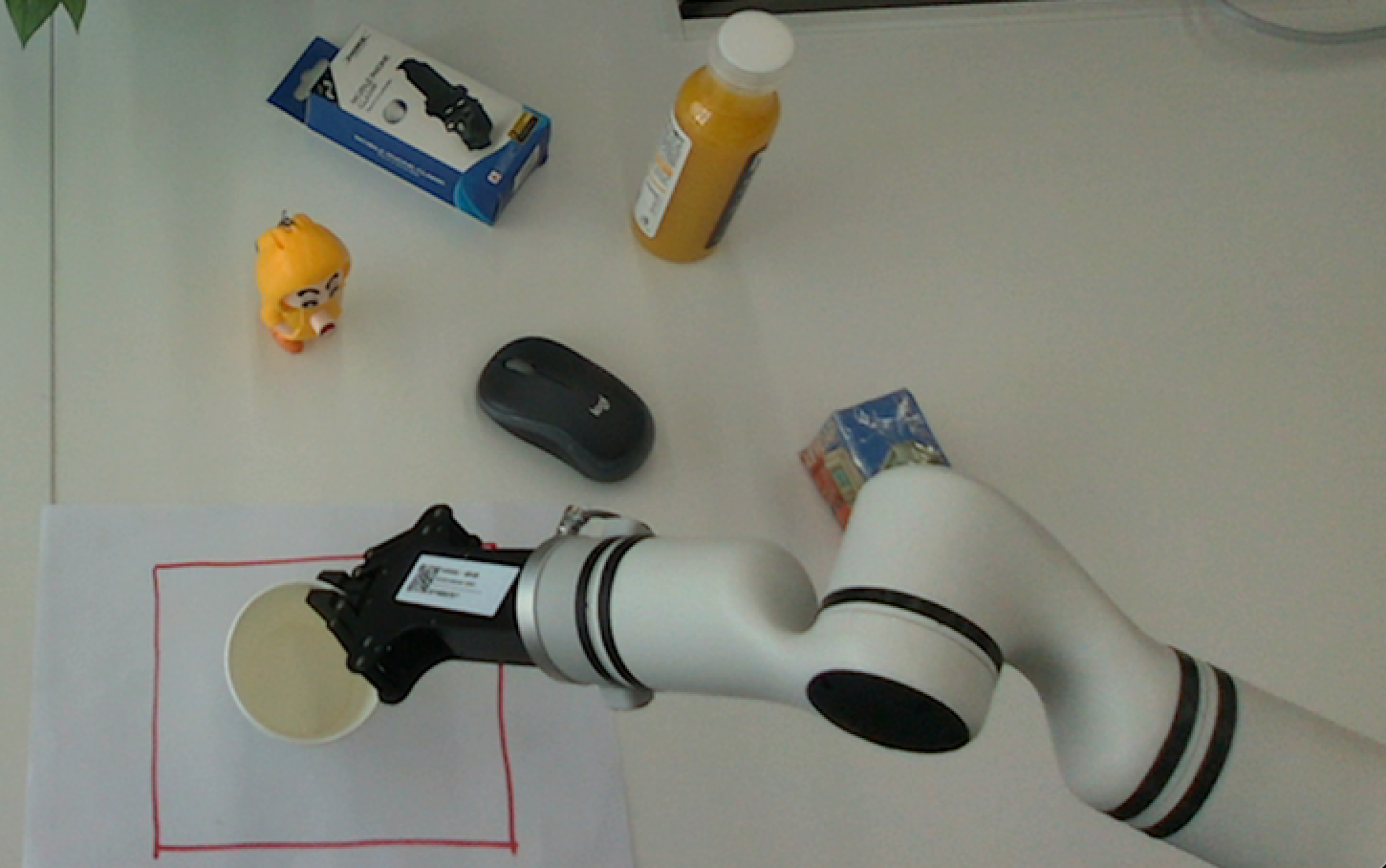}
      \label{fig:1f}
  \end{subfigure}
  
  \begin{subfigure}[b]{0.12\textwidth}
      \includegraphics[width=\textwidth]{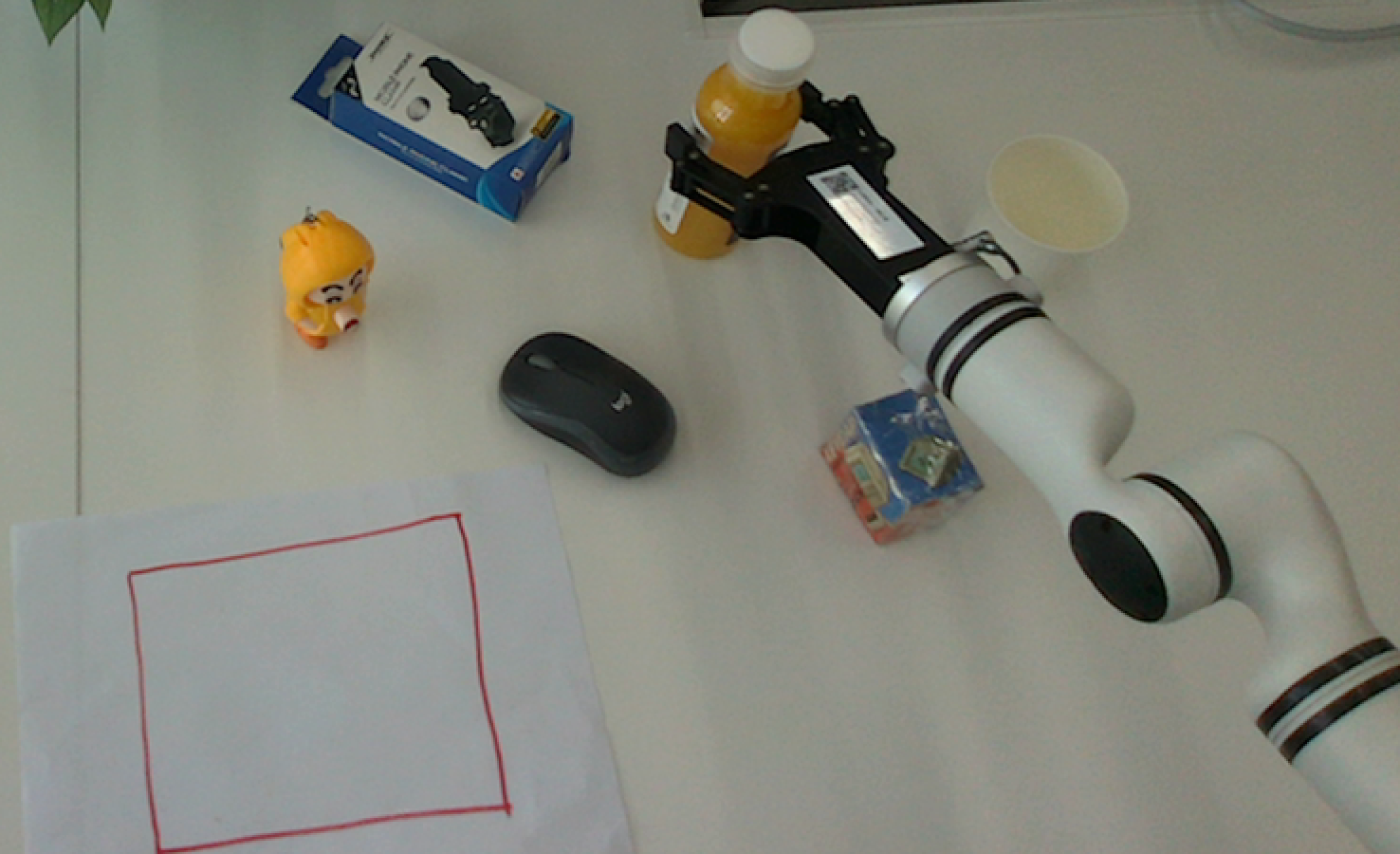}
      \label{fig:1g}
  \end{subfigure}
  ~
  \begin{subfigure}[b]{0.12\textwidth}
      \includegraphics[width=\textwidth]{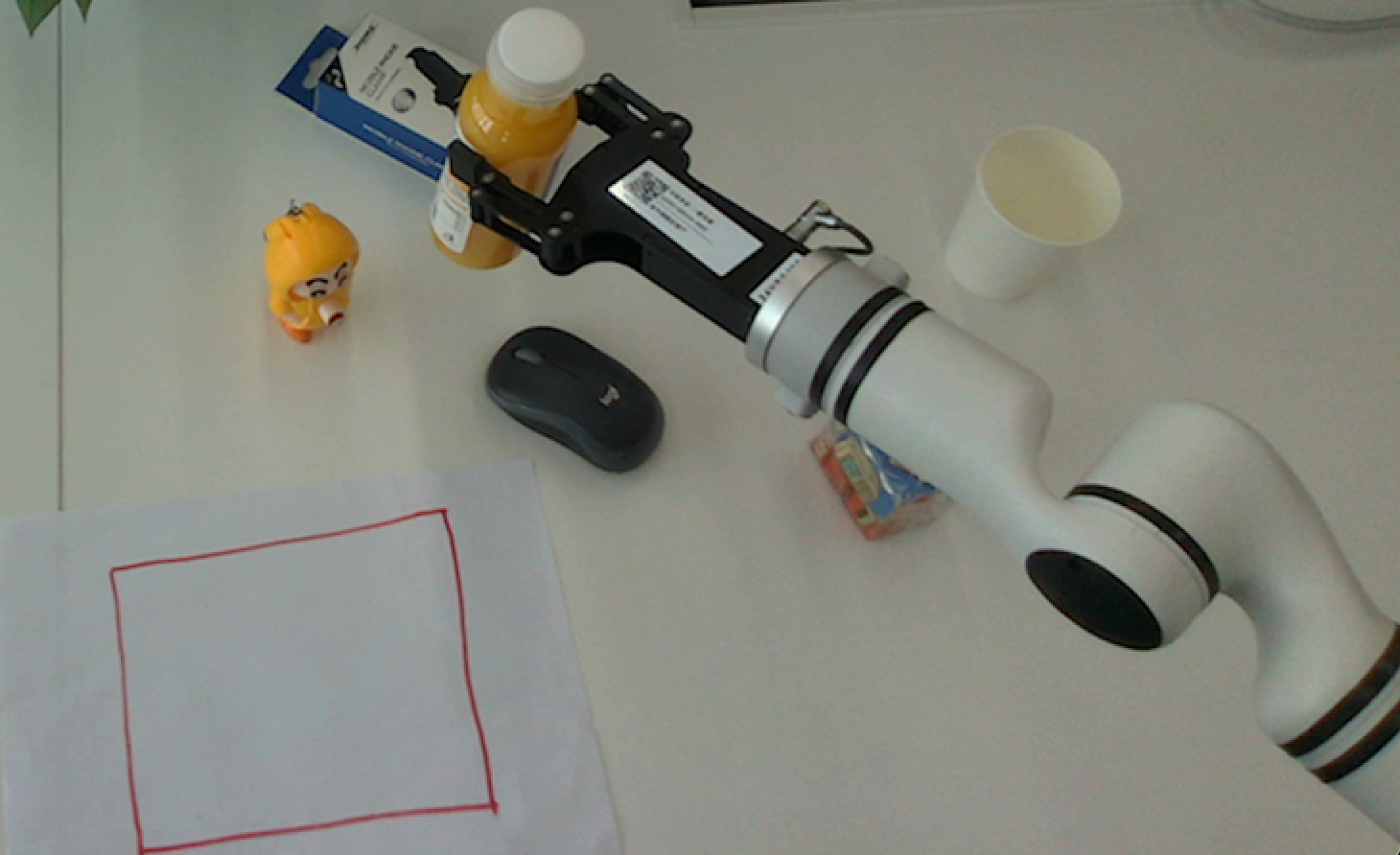}
      \label{fig:1h}
  \end{subfigure}
  ~
  \begin{subfigure}[b]{0.12\textwidth}
      \includegraphics[width=\textwidth]{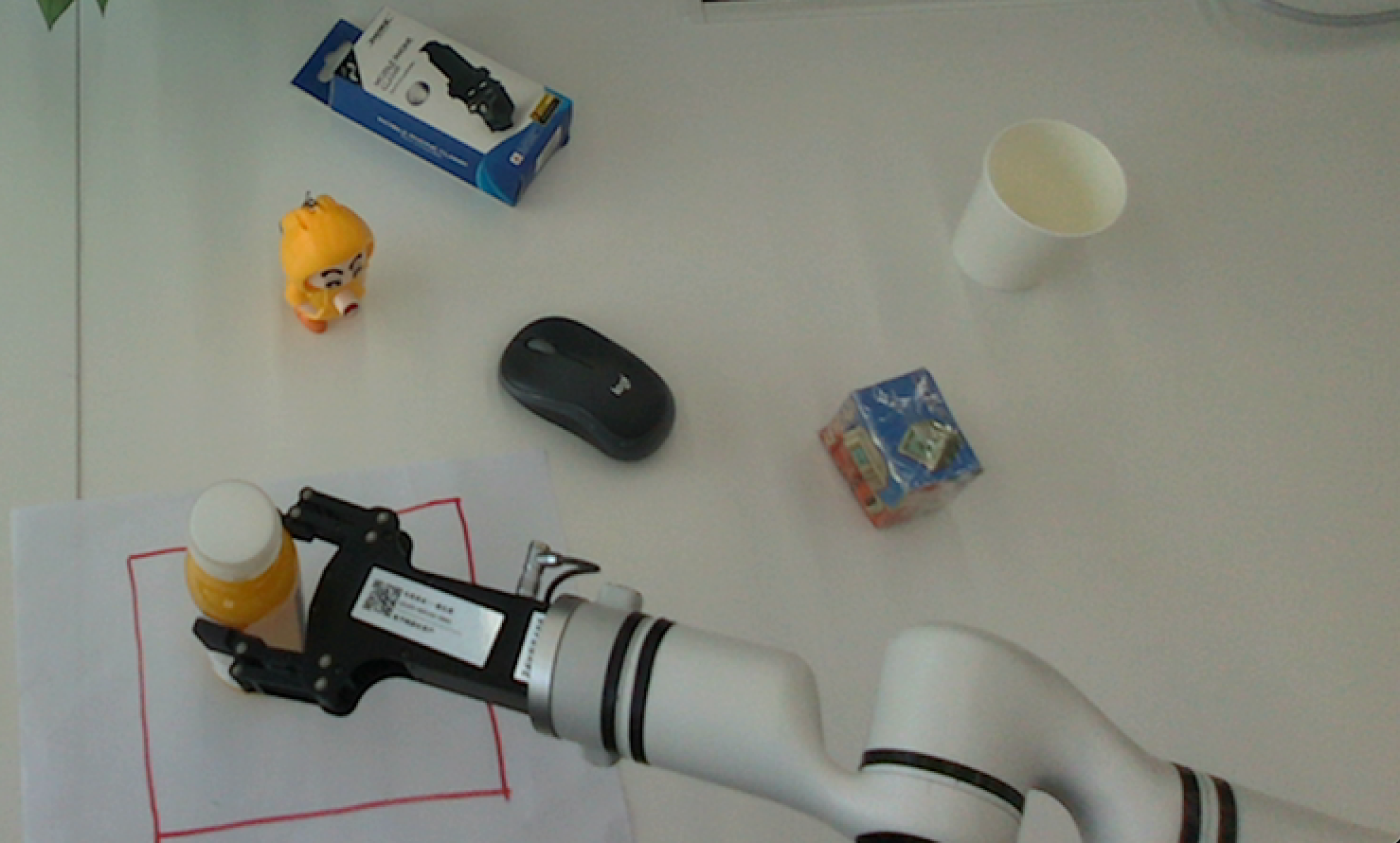}
      \label{fig:1i}
  \end{subfigure}

  \caption{Real-world experiment results for CF-SDP. The first raw shows the $pnp-box$ task, the second row shows the $pnp-cup$ task, and the last raw shows the $pnp-bottle$ task.}
  \label{fig:cf-sdp}
\end{figure}

We design three tasks to evaluate the real-world performance of CF-SDP: $pnp-box$, $pnp-bottle$, and $pnp-cup$, using 0, 1, and 2 to set task-conditions for each task. We collect 50 demonstrations for each task, and during the collection process, the positions of objects were randomly set within a small range. The CF-SDP model is trained for 3000 epochs using all 150 demonstrations. We do not select checkpoints, only the final checkpoint is used for evaluation. In real-world experiments, focusing on two key aspects: success rate and time efficiency. As illustrated in Fig.\ref{fig:cf-sdp}, the CF-SDP can generate actions based on specific task conditions, enable robot to complete specified tasks. Performance is evaluated over 20 eposides for all three tasks. The three tasks—$pnp-box$, $pnp-bottle$, and $pnp-cup$—achieve success rates of approximately $75\%$, $60\%$, and $80\%$, respectively, with 10 denoising diffusion steps. The inference speed of 10-steps, 5-steps, 3-steps, 2-steps, 1-step is 158ms, 81ms, 57ms, 34ms, 18ms, respectively.

\section{Conclusion}
\label{sec:concludion}

We propose the CF-SDP to enhance the efficiency and flexibility of policy generation in imitation learning. By integrating classifier-free guidance and shortcut-based acceleration, our approach significantly reduces the iterative denoising steps, achieving nearly 5× acceleration compared to conventional DDIM-based methods. This enables real-time robot control while maintaining high task performance.

Furthermore, we extend diffusion modeling to the SO(3) manifold, where we formulate the forward and reverse processes in its tangent space using an isotropic Gaussian distribution. This improves rotational estimation, ensuring greater accuracy and stability in motion control.

Evaluations on Robotwin simulator and real scenarios, demonstrate the effectiveness of our approach. Our method not only accelerates diffusion-based policy but also enhances task adaptability, making it a promising solution for embodied robotic manipulation. Future work will explore optimizations in multi-task generalization and real-world deployment. In addition, the results in Table.\ref{tab:performance} show that the optimal sampling steps varies significantly across different tasks. Since the DP3 algorithm can not be reproduced the scaling law, we leave the analysis of optimal step selection to future work, where we will explore it in the context of VLA algorithms with diffusion policies.
{
    \small
    \bibliographystyle{ieeenat_fullname}
    \bibliography{main}
}


\end{document}